# Politeness Counts: Perceptions of Peacekeeping Robots

O. Inbar, J. Meyer (SM), Dept. of Industrial Engineering, Tel Aviv University, Tel Aviv, Israel

*Abstract*—The 'intuitive' trust people feel when encountering robots in public spaces is a key determinant of their willingness to cooperate with these robots. We conducted four experiments to study this topic in the context of peacekeeping robots. Participants viewed scenarios, presented as static images or animations, involving a robot or a human guard performing an access-control task. The guards interacted more or less politely with younger and older male and female people. Our results show strong effects of the guard's politeness. Age and sex of the people interacting with the guard had no significant effect on participants' impressions of its attributes. There were no differences between responses to robot and human guards. This study advances the notion that politeness is a crucial determinant of people's perception of peacekeeping robots.

## I. Introduction

Knightscope is a security robot that navigates autonomously in crowded places, such as shopping centers, interacting with people using sensors and cameras. With no offensive capabilities, it serves as a security guard that scans license plates, detects specific noises from the environment, and follows signals from mobile phone devices. How should one design such a robot, and how should it interact with humans? Should it be friendly or intimidating or some combination of the two?

Indeed, the acceptance of robots in our daily environment is a critical element in human-robot interaction research, e.g., [1], [2]. Six attributes of robot acceptance are representational, physical, behavioral, functional, social, and cultural [3]. Two of these attributes - functional and social - appear to be essential for effective human-robot interaction, and they are both related to users' trust in the robot [3], [4].

Trust is essential in human-machine and human-robot interaction [5], [6]. It makes it possible for people to form attitudes about possible future agent actions [7], and contribute to a more effective interaction [8]. Trust can be a measure of the quality of human-robot interaction, and it is a valid indicator of the robots' functional and social acceptance [3], [9].

Trust, in the context of automation, generally refers to aspects of a person's attitude towards the automation. As with other attitudes, it may have affective, behavioral or cognitive components, and different views of trust in automation stressed each of these components [10]. It is a reflection of the perceived predictability, reliability, and capability of automated systems, such as those for automotive safety, healthcare and industrial devices [5]. Trust is influenced by individual differences [11], prior information and past experiences [12], similar to human-human trust that is quickly generated and then fine-tuned over time [13]. Three key components are essential to inspiring trust: a human, a system, and two-way communication between the first two that provides feedback to the human [14].

Inspiring trust in robots is complicated, as robots are typically complex systems with judgment and actions that are neither fully comprehensible nor fully transparent to humans interacting with them [15]. Another challenge is due to the effect of time: most research on human-robot interaction examines repeated interactions, which enable people to adjust their level of trust during the experiment(s) [4].

In this paper, we inspect the use of robots as autonomous peacekeepers in a civilian environment. In this context, robots have the obvious advantage that they raise the physical safety of peacekeeping personnel by minimizing human-to-human encounters during conflict [16]. Moreover, equipping robots with non-lethal weapons enhances their value, in addition to reducing moral and ethical concerns arising from the use of lethal weapons [17]. The advantages for personnel, in addition to the ability to equip robots with non-lethal weapons, is leading to a rapid increase in the use of peacekeeping robots in military and police operations [9, 18].

Designing peacekeeping robots that interact with civilians involves several considerations. First, a human interacting with a peacekeeping robot does not operate the robot, as in standard robotic systems [19]. Rather, he or she is a "bystander", a person who "does not explicitly interact with a robot but needs some model of robot behavior to understand the consequences of the robot's actions" [20]. A bystander has little or no control over the robot's actions, lacks information regarding the robot's

---

Manuscript submitted April 21 2017. This research was funded by the U.S. Office of Naval Research – Global (ONR-G) under grant N62909-14-1-N180.

The authors are with the Department of Industrial Engineering, Tel Aviv University, Tel Aviv 6997801, Israel (email: ohadinbar@tauex.tau.ac.il; jmeyer@tau.ac.il).

Color versions of one or more of the figures in this paper are available online at http://ieeexplore.ieee.org.

intentions, goals and expected behavior, and is generally passive [21], [22]. As a result, a comprehensive interface that involves both social and cognitive interactions between humans and robots is necessary [23], [24]. One can possibly design robots to express social characteristics associated with humans, such as moods and emotions [25], using natural cues (e.g. gestures) and exhibiting distinctive personality. This may help to achieve transparency, so that a system's action, or the intention of an action, is apparent to human operators and/or observers [26].

Second, in traditional automation, a well-calibrated level of trust requires numerous repetitions [27]. However, in the context of peacekeeping robots, interactions may be occasional and infrequent, limiting the humans' ability to adjust their trust in robots. This issue is even more crucial with first time encounters that require humans to form initial trust almost immediately. Consequently, initial trust in these cases results, in part, from the robot's characteristics, such as its appearance and behavior [14]. In particular, the robot's appearance (shape, structure, aesthetics) plays an important role in establishing social expectations [28], as it affects people's interaction with the robot [23], and therefore should match the robot's intended function [29]. Specifically, for robotic peacekeepers, the robot's appearance should convey attributes such as agency (apparent ability of the robot to use weapons as protective measures) and authority (apparent power to act in a manner representative of a peacekeeping force) [30].

Third, in many of these interactions the robot faces not just one individual, but rather a group of people who jointly interact with the robot, e.g., in markets or protests [31, 32]. As a result, the individual's trust derives not only from the personal experience when interacting with the robot, but also from observing the robot interact with other people, which can also lead to more complex crowd reactions [33]. Other sources of information can also contribute to the level of trust, such as media, word-of-mouth and more.

Fourth, the interaction takes place in a dynamic environment [6], involving possible physical interaction between robots and humans, for instance in access control to buildings or market checkpoint scenarios [32]. Human diversity also adds to the complexity, as age, sex, level of education and technology literacy may influence the acceptance of robotic technology and the trust in such systems [24]. In this context, cultural (and especially national) differences should be taken into consideration [34], as robots can be deployed at various locations and in different cultures around the world, and since cultural factors, such as uncertainty avoidance and individualism/collectivism, influence trust [26].

Social skills are important in human-human social interactions, where various cues contribute to the perceived trust people have in other people [5]. They are also essential assets for robots interacting and communicating with people [35, 36, 37], as humans are inclined to interpret computer behaviors similarly to human-human interactions [7, 38], with computers serving as 'social actors', as proposed by the CASA (Computers Are Social Actors) theory [39, 40, 41]. Among social skills, politeness is a key element [42, 43]. [44] described politeness, based on the work of Brown and Levinson [45], as "one means by which we convey, interpret, maintain and alter social relationships". While there is no generally accepted definition of politeness, its meaning usually refers to something like "showing good manners and consideration for other people" (Oxford Advanced Learner's Dictionary). Goffman [46] stated that politeness fulfils an important social function, as it makes possible communication between potentially aggressive parties. Considering that peacekeeping robots are almost by definition potentially aggressive, it may be particularly relevant in the interaction with them.

Etiquette is a set of unwritten codes, by which we signal politeness in a specific social context according to cultural norms. It makes use of verbal, physical, gestural, and even more primitive modes of interaction [7, 44]. Etiquette also helps to identify individuals as members of trustworthy groups [36]. Proper etiquette of robotic systems can have considerable influence on people's acceptance and the optimal use of these systems, and more importantly in this context, their trust in these robotic systems [36], [11]. Robotic etiquette can also influence people's reactions towards robots, as people tend to respond to automated systems in a manner similar to human-human social interactions [47]. Thus, peacekeeping robots interacting with civilians according to the anticipated etiquette (e.g., treating elderly people, children, and disabled people more politely) may be considered correct behavior, and it may generate more trust and cooperation from the population.

This paper expands the preliminary study presented in [48], which showed that participants evaluated a robot's behavior only as a function of the politeness of the robot, while disregarding other factors, such as the age and sex of the person with whom the robot interacted. The paper is structured as follows: In Study 1 we replicated the preliminary study of Inbar and Meyer [48] with a larger number of participants from a different country and with a different data collection method. In Study 2, we evaluated a range of levels of politeness. In Study 3 we used four levels of politeness and measured participants' reactions to a robot guard and to a human guard. In Study 4, we used animated clips instead of the static images used in the previous studies. We end the paper with some conclusions, drawn from the set of studies.

## II. STUDY 1

The original study of Inbar and Meyer [48] was limited both in the number of participants (30) and in the experimental design, as it was conducted in a classroom, using printed stimuli. In Study 1 we replicated their study with a larger number of participants, using online data collection, with the goal of reproducing similar results with a more robust design.

*1) Method*

  *a) Participants*

We recruited participants for this study, using Amazon's Mechanical Turk (mTurk). Of the 99 participants that completed the study, 54 were male, and 51 were younger than 34 years old. All participants were US residents, and were paid a fixed amount, based on mTurk tariff.

*b) Experimental Design*

As in the previous study, we manipulated three independent variables: the age of the individual presented in each situation – young (~20) or old (~70); the sex of the individual (male or female); the extent of politeness expressed by the robot's dialog while conversing with the individual (polite or impolite). A fourth variable we added to the analyses was the participant's sex (male or female).

The scenario was again an "access control" situation, in which a robot is in charge of inspecting people entering a building, deciding on the correct reaction: inspection, blocking or passing. The robot's interaction with the people is represented as a 'call out' text message. We generated the scenes, using a picture of a real-looking entrance to a building and pictures of real people (male and female, old and young). Similar to the previous study, we decided to use the same humanoid robot model for this study (which we considered vaguely intimidating, due to its muscular appearance and the use of metal construction). Although a robot's appearance plays an important role in people's reaction to it [49], for the purpose of this study, we chose not to manipulate this (rather complex) variable. For the same reason, we also chose to keep a 'neutral' facial expression of the robot.

Participants received a link to the survey, which we conducted using the Qualtrics survey platform. Participants first filled-in a consent form, and subsequently saw eight scenarios (2 age x 2 sex x 2 politeness). Figure 1 depicts one of these conditions - a young female interacting with a polite robot. Figure 2 presents all eight conditions.

To minimize order and learning effects, we randomized the order of the scenarios individually for each participant. Participants received a brief introduction, stating that this study deals with peacekeeping robots performing access-control duty at the entrance to an office building.

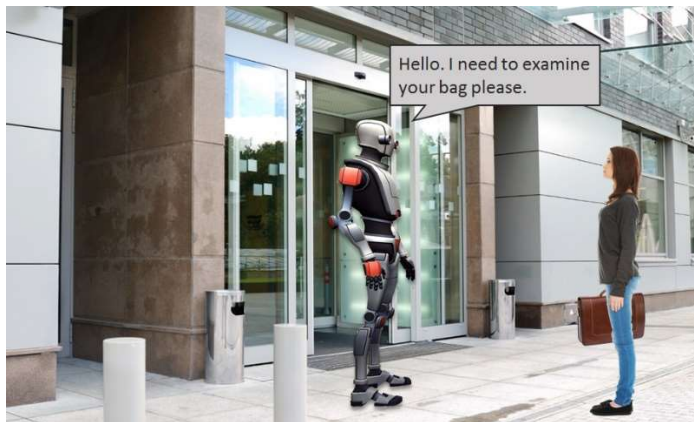

Fig. 1. Young woman, polite robot behavior

Participants observed the scenes one at a time, as a third-person evaluating the interaction between a robot and a human, and then answered the following questions for each condition (with no time limit):

1. How intimidating is the robot? (Dependent variable: Intimidating)
2. How fair is the robot? (Dependent variable: Fairness)
3. How friendly is the robot? (Dependent variable: Friendly)

We collected the answers to these questions using a 5-point Likert scale, with 1 being "Extremely not" and 5 "Extremely yes". We also asked about the respondent's sex.

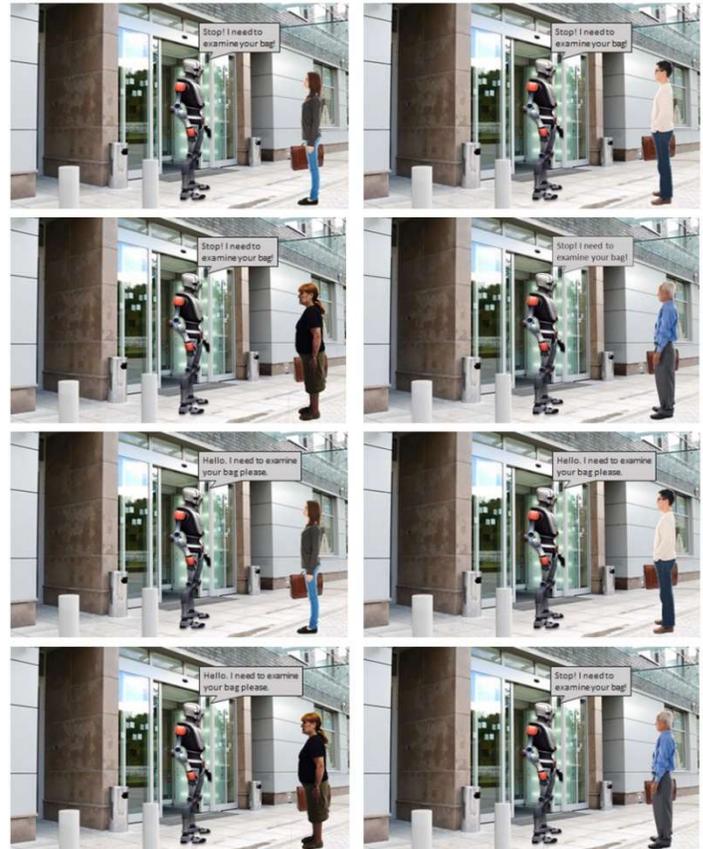

Fig. 2 The eight conditions, resulting from all combinations of: young and old, male and female, polite and impolite robot.

*2) Results*

We analyzed the three dependent variables with separate mixed (within-between) Analyses of Variance (ANOVAs) for each dependent variable, with the respondent's sex (male, female) as a between-subject variable, and the robot politeness (polite, impolite), the age of the person interacting with the robot (younger, older), and the sex of the person interacting with the robot (male, female) as within-subject independent variables. Table I shows the results of the analyses for the three dependent variables for the effect of the robot politeness. None of the other main effects or two-way interactions was significant.

As in the original study, the strong effects of the robot's politeness show that it is the first and most important determinant of our participants' view of the robot. It also makes sense that when a less polite robot simply issues commands, people see it as more threatening than a robot that asks people to comply with its orders in a politer way.

TABLE I. Effects of the robot's politeness on the three dependent variables (all effects were significant with p<.001). Shown are the F values and the means (and standard errors) for the polite and impolite robot.

|  | $F_{(1,95)}$ | Partial $\eta^2$ | Polite (means/SE) | Impolite (means/SE) |
| --- | --- | --- | --- | --- |
| Intimidating | 94.99 | .5 | 3.31 (0.11) | 4.27 (0.09) |
| Fair | 77.02 | .45 | 3.2 (0.1) | 2.41 (0.1) |
| Friendly | 309.0 | .77 | 3.52 (0.1) | 1.66 (0.08) |

However, somewhat less predictably, judgments of the robot's fairness also depended on the degree to which the robot acted politely. This means that respondents did not consider fairness as the degree to which the robot treats all people it encounters equally, but rather people based their assessments on the politeness of the robot. Unsurprisingly, people also evaluated the friendliness of the robot based on its politeness.

These results resemble those of our original study. Again, the results show that a robot is considered more intimidating when displaying "impolite" behavior. We also demonstrated again that participants' attitudes were not affected by expected etiquette, as they were indifferent to the robots' politeness toward specific groups of people, towards whom the robots might have been expected to act more politely (e.g., the elderly or women).

*3) Discussion*

While these results are comparable to our initial study, they are still somewhat surprising: We expected that conventional etiquette would lead participants to consider robots as less polite when they treat an older lady in an unfriendly way, compared to treating a younger man in the same manner. In other words, we expected participants to judge the robot's impolite behavior more harshly when being impolite towards women (young and old) and older people, as we would have assumed they would with a human guard behaving in the same manner. A possible explanation for these results can be that the effect of the robot's impolite behavior was so overwhelming that it dominated participants' attitude, and dominated other factors, such as age and sex of the person with whom the robot interacted. Another explanation is that undifferentiated politeness is indeed a key factor, shaping people's perception of peacekeeping robots, affecting the degree to which they view their actions as correct and fair.

### III. STUDY 2

The results of Study 1 left us questioning whether the use of only two levels of politeness – polite and impolite – could explain participants' lack of concern about social etiquette. In Study 2 we sought to examine this question, and to create a more refined scale of politeness.

Brown and Levinson, in one of the most influential works on politeness, suggested the theory of "face" as the foundation of the theory of politeness, with "face" being the public image a person presents to others and attempts to protect [45]. Undermining this image is regarded as a "face-threatening act" (FTA), and it is the role of politeness to soften such threats. Brown and Levinson defined four main communication strategies, based on negative or positive face-threatening acts, and on the role in the conversation - speaker or hearer [45]. Of these four strategies, we focused on the negative face-threatening act, which occurs when freedom of choice and action are prevented, as in the access-control scenarios used in this study. We further focused on the implications to the hearer – the human in our case – interacting with the speaker – the robot. This communication strategy uses orders, requests, suggestions, advice and even threats to put pressure on the hearer to either perform or not perform an act.

Blum-Kulka [50] defined politeness in the context of requests as the interactional balance between two needs: The need for pragmatic clarity and the need to avoid coerciveness. This theory is based on the typology of request patterns that includes nine categories of request patterns, from the least polite, termed "Mood Derivable" (e.g. "Move your car") to the most polite, termed "Mild Hints" (e.g. "we don't want any crowding" – as a request to move the car). We referred to these categories, when creating the politeness scale for our study.

*1) Method*

*a)   Participants*

We recruited participants again through Amazon's Mechanical Turk (mTurk). 118 participants completed the study: 57 of them were male, and 61 were female. All participants were US residents, and were paid a fixed amount, based on mTurk tariff.

*b)   Experimental Design*

Based on [50], we chose a set of six requests, consisting of all combinations of two opening words – "Hello" and "Stop" – and three different types of requests: a most polite phrase ("Would you mind letting me inspect your bag?"), an intermediately polite phrase ("Could I inspect your bag?"), and a least polite phrase ("Give me your bag for inspection"). The combination of the two opening words and the three phrases created six different requests (e.g. "Hello. Would you mind letting me inspect your bag?"), for which we wanted to measure the perceived politeness.

We conducted a study to evaluate the politeness ratings for the six different requests, using Qualtrics as a survey tool. Respondents rated each sentence on a scale from 1 (very impolite) to 5 (very polite).

Results

We analyzed the responses with a three-way ANOVA with the respondent's sex (male or female) as a between-subjects variable and the opening word ("stop" or "hello") and the three phrases as within-subjects variables. We randomized the presentation order of the six statements individually for each respondent.

There was a significant main effect of the opening word, $F(1, 116) = 797.72$, $p<.001$, partial $\eta^2=.87$, with "Hello" being rated significantly more polite than "Stop". Also, there was a significant main-effect of the phrase, $F(2, 232)= 551.27$, $p<.001$, partial $\eta^2=.83$, with higher ratings for the more polite phrases than for the less polite ones. The interaction between the opening word and the politeness of the phrase was also significant, $F(2, 232)=52.47$, $p<.001$, partial $\eta^2=.31$. Table II shows the mean politeness ratings and standard errors for the different combinations.

TABLE II Mean politeness ratings for the six combinations of opening word and phrase (mean and standard error)

|  | Would you mind letting me inspect your bag? | Could I inspect your bag? | Give me your bag for inspection |
|---|---|---|---|
| Stop | 2.52 (0.085) | 2.23 (0.078) | 1.17 (0.046) |
| Hello | 4.62 (0.057) | 3.86 (0.078) | 2.16 (0.081) |

### a) Discussion

The results of this study provided us with a range of levels of politeness, based on the six statements we evaluated. At one end of the scale, the opening word "Stop" was mostly associated with an impolite request, while at the other end of the scale, the opening word "Hello" was mostly associated with a polite request. We used this scale in Study 3 to generate four levels of politeness of either a human or a robot guard in a similar access control scenario.

## IV. STUDY 3

The results of Study 1 brought about another question: were people's reactions to the guard influenced by the fact that it was actually a robot? In other words – would their reaction to a human guard be different?

In this study we replicated Study 1 with two important additions: First, we presented four levels of politeness, based on the results of Study 2 and Blum-Kulka's [50] rankings of the directness of an interaction: "Hello. Would you mind letting me inspect your bag?" (Most polite, with a mean politeness rating of 4.62), "Stop. Would you mind letting me inspect your bag?" (Intermediate polite, with a mean politeness rating of 2.52), "Hello. Give me your bag for inspection." (Intermediate impolite, with a mean politeness rating of 2.16), and "Stop! Give me your bag for inspection." (Most impolite, with a mean politeness rating of 1.17). Second, we presented the scenes with both a robot and a human guard. This allowed us to test for possible differences between the responses to a human and a robotic guard. In particular, we aimed to determine whether more or less polite behavior towards people with different demographic characteristics will elicit different evaluations if a human or a robot guard displayed the behavior.

### 1) Method

#### a) Participants

We recruited 101 participants for this study again with Amazon's Mechanical Turk (mTurk) (57 males and younger than 35 years). All participants were US residents, and were paid a fixed amount, based on mTurk tariff.

#### b) Experimental Design

We manipulated four independent variables: the age of the individual presented in each situation – young (approximately in his or her 20s) or old (approximately in his or her 70s); the sex of the individual (male or female); the type of guard (robot or human), and the degree of politeness expressed by the guard's behavior while conversing with the individual (four levels). The participant's sex was a fifth variable in the analyses.

Similar to Study 1, the scenario was an "access control" situation, in which a guard inspects people entering a building. We used the same humanoid robot model, the same location and the same people interacting with the robot as in Study 1. For the human guard condition, we used the image of a human guard, purchased from a stock photo service.

Participants received a link to the survey on the Qualtrics survey platform. Participants saw one of two conditions, robot guard or human guard. Each participant evaluated 16 scenes, representing all combinations of age (2), sex (2) and politeness (4). Figure 3 presents one example of these conditions, of a young female interacting with a polite human guard.

To minimize order and learning effects, we randomized the order of the conditions individually for each participant. Participants received a brief introduction, stating that this study deals with guards performing access-control duty at the entrance to an office building.

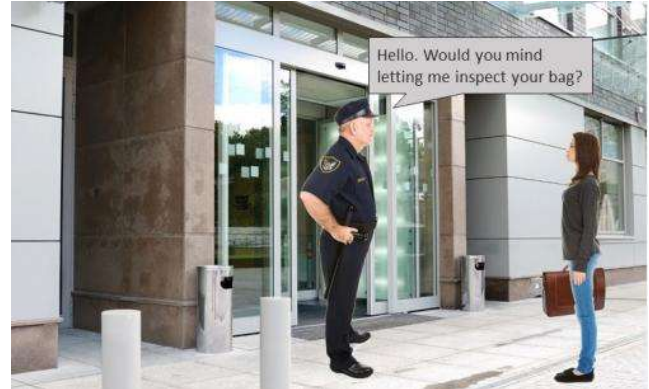

Figure 1 - Young woman, polite human guard behavior

We directed participants to examine the conditions one at a time, and to answer the following questions for each condition (with no time limit):

1. How intimidating is the guard? (Dependent variable: Intimidating)
2. How fair is the guard? (Dependent variable: Fairness)
3. How friendly is the guard? (Dependent variable: Friendly)

We collected the answers to these questions using a 5-point Likert scale, with 1 being "Extremely not" and 5 "Extremely yes". We also asked about the respondent's sex.

### B. Results

We analyzed the three dependent variables with mixed within-between subject ANOVAs with the type of guard (human, robot) and the respondent's sex as between-subject variables, and the robot behavior (four levels of politeness), the age of the person interacting with the robot (younger, older) and the sex of the person interacting with the robot (male, female) as within-subject independent variables.

As in Study 1, the guard's politeness is the first and most important determinant of our participants' view of the guard. Table III shows the results of the analyses for the three dependent variables for the effect of the guard politeness. None of the other main effects or two-way interactions was significant.

TABLE III Effects of the guard's politeness on the three dependent variables. Shown are the Degrees of Freedom (df), the F values and the means (and standard error) for the different levels of politeness. All results are shown with Greenhouse-Geysser corrections for violation of the sphericity assumption and all were significant at p<.0001

|  | df | F | Partial $\eta^2$ | Most Polite | Med Polite | Med Impolite | Most Impolite |
|---|---|---|---|---|---|---|---|
| Intimidating | 1.96, 559.1 | 133.98 | .6 | 2.49 (0.11) | 3.44 (0.09) | 3.50 (0.09) | 4.25 (0.09) |
| Fair | 2.04, 557.6 | 117.92 | .57 | 3.80 (0.11) | 2.74 (0.10) | 3.05 (0.1) | 2.07 (0.1) |
| Friendly | 2.95, 1178.8 | 276.1 | .76 | 4.03 (0.08) | 2.64 (0.09) | 2.81 (0.1) | 1.48 (0.05) |

Similar to Study 1, participants in this study judged the guard's behavior based on his level of politeness. Politeness affected intimidation, with less polite behavior perceived as more intimidating. The guard's politeness also affected perceived fairness and friendliness, with politer behavior generating friendlier and fairer impressions. None of the other main effects or interactions were significant.

*C. Discussion*

The results of Study 3 echo those of Study 1. Although we used a more refined politeness scale, there was no significant difference in the evaluations of robots and humans serving the same access-control guarding function. Participants' perceptions of the guards (whether they were robots or humans) were unaffected by the guards' behavior toward specific groups of people, such as being more polite towards the elderly or women, as perhaps prescribed by etiquette. The finding that people's expectations from robot guards are similar to those from human guards resembles the results in research on devices as social agents [51]. People are generally willing to treat robotic devices as social actors, as long as they allow the humans to form a stable behavioral model of the robot, and it is possible to predict the robot's behavior from observable attributes of the participant [52].

## V. STUDY 4

The results of Study 1 and Study 3 were consistent. However, they may be due to our use of static images to present the access control situation. Properties of the interaction, such as the characteristics of the human interacting with the robot (sex, age) or the guard, may affect a viewer less with static images than with animations. An animated video clip may enhance some demographic characteristics, such as the pace of walking or climbing stairs, which in turn, may, for example, help participants relate more to the age of the depicted people. Some support for this argument can be found in the domain of learning and instruction, where a meta-analysis [53] shows a clear advantage of animations compared to static pictures in the instructional domain. Finding different patterns of results for static images and dynamic clips would have methodological implications, namely that one has to prepare some kind of dynamic depiction of interactions to study the responses to different types of communication.

To address this matter, we replicated Study 3 with dynamic stimuli to make the interaction between the guard and the human more vivid. Participants saw a human approaching a guard, followed by the guard's request from the human. The presentation through animations may enliven the human as an individual, which may cause participants to consider the demographic characteristics of the human with whom the robot interacts more or less politely. In this experiment, we only used the two extreme phrases from Study 3, because the presentation of the animated stimuli took time, and we could only present a limited number of conditions in a reasonably long experiment. We also added another dependent variable – appropriateness of operation – to judge the perceived correctness of the functioning of the guard.

*A. Method*
*1) Participants*

We again recruited participants for this study with Amazon's Mechanical Turk (mTurk). 213 participants completed the study: 126 of them were male and 127 were younger than 34 years old.

*2) Experimental Design*

The experimental design was very similar to Study 3. We manipulated the same four independent variables: the age of the individual presented in each situation (young, ~20yo or old, ~70yo); the sex of the individual (male or female); the type of guard (robot or human), the extent of politeness expressed by the guard's behavior while conversing with the individual (polite or impolite, using the two extreme politeness levels from Study 3).

A similar "access control" situation was used, in which a guard inspects people entering a building. For this study we used Unity3D software to create the animations and presented the guard's request in a text box at the bottom of the frame. Similar to Study 1 and Study 3, we decided to use a humanoid robot model and a standard model of a male human guard. We sent participants a link to the Qualtrics survey. In it, they saw clips from one of two conditions: a robot guard or a human guard. For each set, participants saw each of the 8 conditions (2 age x 2 sex x 2 behaviors). The order of the conditions was randomized for each participant, and participants received a brief introduction before starting the study.

Each animation showed a person crossing the street in the direction of an office building, climbing a few stairs, and meeting a guard. The guard then communicates with the approaching person. The communication appears as sub-titles (to avoid the possible effects of intonation and other non-verbal features of speech). Animations were 12 seconds long and started playing automatically after the page loaded. It was only possible to move to the next animation after the previous animation finished playing.

Figure 4 presents a sequence of thumbnails representing one animation, of a young female interacting with a polite robot (the thumbnail numbers present the progress in the animation).

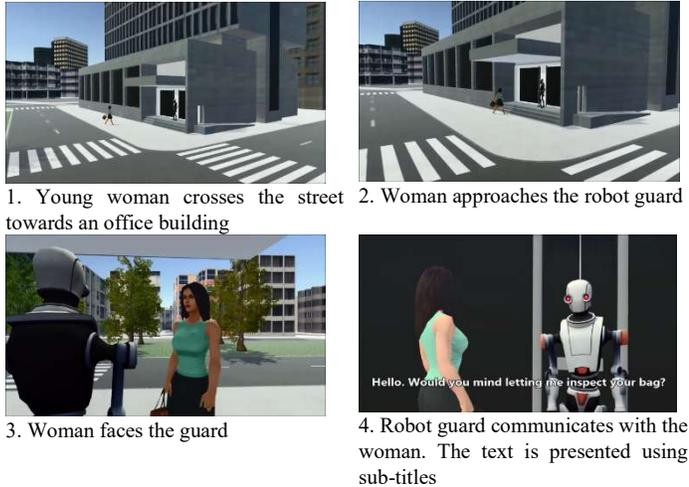

1. Young woman crosses the street towards an office building
2. Woman approaches the robot guard
3. Woman faces the guard
4. Robot guard communicates with the woman. The text is presented using sub-titles

Figure 2 - Thumbnails of an animation of a young female interacting with a polite robot

After each animation, participants indicated their level of agreement with four statements (with no time limit):

1. The guard is intimidating (dependent variable: Intimidating)
2. The guard is fair (dependent variable: Fairness)
3. The guard is friendly (dependent variable: Friendly)
4. The guard's behavior is appropriate

We collected the answers to these statements using a 5-point Likert scale, with 1 being "Extremely not" and 5 "Extremely yes". We also asked about the respondent's sex.

### B. Results

We analyzed the four dependent variables with mixed within-between subject ANOVAs with the type of guard (human, robot), the respondent's sex and age (young, older) as between-subject variables, and the robot politeness (polite, impolite), the age of the person interacting with the robot (younger, older) and the sex of the person interacting with the robot as within-subject independent variables.

As in the previous studies, the robot's politeness is the first and most important determinant of our participants' view of the robot. Table IV details the results of the analyses for the four dependent variables for the effect of the robot attitude. None of the other main effects or two-way interactions was significant.

Similar to Study 1, participants in this study judged the guard's behavior, based on his level of politeness. Politeness again affected intimidation, with less polite behavior seeming more intimidating. Politeness also affected judgments of fairness, friendliness and appropriateness, with more polite behavior generating friendlier and fairer impressions.

TABLE IV Effects of the robot's politeness on the three dependent variables (all effects were significant with p<.001). Shown are the F values and the means (and standard error) for the polite and impolite robot.

|  | $F_{(1,205)}$ | Partial $\eta^2$ | Polite Mean (std. err.) | Impolite Mean (std. err.) |
|---|---|---|---|---|
| Intimidating | 497.6 | .71 | 2.20 (0.08) | 4.14 (0.06) |
| Fair | 521.11 | .72 | 4.17 (0.05) | 2.31 (0.07) |
| Friendly | 1568.51 | .88 | 4.27 (0.05) | 1.57 (0.05) |
| Appropriate | 619.06 | .75 | 4.25 (0.05) | 2.11 (0.08) |

### C. Discussion

The results of Study 4 are in line with those of Studies 3 and 1: we see again that participants' assessments of guards' attributes, such as intimidation, fairness, friendliness and appropriateness of behavior, are based mainly on the guards' politeness. Polite guards are considered to be friendlier, fairer and with more appropriate behavior in general. However, polite guards, as expected, are also considered to be less intimidating – a result that can also be considered as less favorable, depending on the image the guard is supposed to present.

Once again, we witness no significant difference between robots and human guards, and no effect of common etiquette on participant's assessment of the guards' conduct. For example, treating elderly people harshly, which would normally be regarded as a ruthless behavior, did not result in participants 'punishing' the guard (with lower ratings for appropriateness of behavior) for not complying with accepted etiquette.

It also appears that replacing static pictures with animated clips had little or no influence on the affect generated toward the humans interacting with the guard, especially women and elderly people.

## VI. FINAL DISCUSSION

Robotic peacekeepers are already a reality: Knightscope, the security robot mentioned at the beginning of this paper is one example. Robo-Guard, a prison patrolling robot in correctional facilities in South Korea, is another. These robots serve as guards, replacing or assisting security guards or police and military personnel, and interacting with local populations in social settings. Robots interacting with humans have an inherent conflict: on the one hand, they need to stick to their missions; on the other hand, they need to obey social rules and behave in a socially acceptable way [52].

Typically, people's first encounter with these law-enforcing robots occurs when they are unprepared and have few or no preconceptions, so behavior toward the robots is determined by the rapid, intuitive trust established during the initial interactions. Imperative to inspiring such trust is the use of appropriate cues, including careful consideration of the appearance and behavior of the robots.

This paper attempts to contribute to the discussion on the social cues required to ensure optimal interactions between peacekeeping robots and humans, and more specifically, the social cues that are apparent through communication style and politeness. In order to explore this matter, we conducted four studies, evaluating how people react to situations in which a more or less polite robot, responsible for access-control, interacts with people of different age-groups and sex.

Overall, our results show a large effect of the robot's politeness on the way the robot is perceived. In all studies, the only factor influencing the participants' perception of the robot's behavior was its manners. Polite robots were perceived as friendlier, fairer and as acting in a more appropriate way. Polite robots were also perceived as less intimidating. There are several additional noteworthy findings from this series of studies:

First, social etiquette seems to matter little when it comes to robot behavior. A more polite approach towards older people, and perhaps women, was not considered as a virtue, and robots that were impolite toward elderly women were not 'penalized' for this behavior. These results are somehow surprising, as we expected that accepted social etiquette would cause participants to anticipate a gentler, more polite approach towards certain segments of the population. These impressions, remarkably, were unaffected by the demographic characteristics of the participant observing the interactions. We had male and female participants with a wide range of ages in our experiments. Hence, this seems to be a robust finding that does not depend on participants' demographics. A possible explanation for these findings is that when the gap between a polite and an impolite guard's behavior is so striking, it overshadows the other differences of age and sex. However, the effect was also evident in Study 3, where guards displayed more refined forms of politeness. While we may have expected differences between human and robot guards in the intermediate levels of politeness, no such effect appeared.

Second, robot guards and human guards received similar assessment in both studies (the first study presented only robot guards). While our study focused on a specific scenario, with a specific communication style, and with a particular humanoid look for the robot, we can cautiously speculate that people's expectations from robots' behavior may be quite similar to their expectations from humans performing similar roles.

Third, our findings seem to be rather stable over geo-locations: the preliminary study [48] was conducted in Israel (using Hebrew for text), while studies 1 to 4 (reported in this paper) were conducted with English speaking mTurk respondents from the U.S.A.. Still, the patterns of results were very similar, and there was no evidence for cultural or language differences. Our findings were also stable over different levels of fidelity and experimental method, as both had no apparent effect on our findings: from paper-and-pencil stimuli in a classroom, followed by an online study using static images, to an online study using animated clips.

Human-robot interaction has benefited from research on 'social robots' interacting with people, mostly in the context of helping them in their daily lives, and assisting people with physical and cognitive disabilities [54]. However, research on military and peacekeeping robots often tends to overlook the importance of conforming to human etiquette, especially for non-lethal robots interacting with civilians, while maintaining compliance and obedience of people with these robots. This paper presents additional evidence that politeness is not a superficial feature, but rather a crucial determinant of people's perception of peacekeeping robots, affecting the extent to which people view the robot's actions as correct, efficient and fair. There was no indication that people might accept a "tough but fair" robot, or that a less polite robot might seem more capable.

There are several limitations to the studies reported in this paper, which could be resolved in future studies: First, using the same robot for all studies had the advantage of keeping this variable constant, but it limits our understanding of effects of the robot shape on the way it is perceived [23]. Using different robots in future studies may help to answer questions related to robot shape. Second, in both static images and animated clips, we used only textual messages to present the guard's instructions. The use of sound and speech may make the scenario more realistic, although it may also add additional variables that one needs to control. Third, although the studies were conducted in two different cultures (the US and Israel), which have distinct characteristics, a more thorough exploration of other cultures may be beneficial to understand additional aspects that relate to language and culture.

Finally, the study employed depictions of encounters, and the question obviously arises how these findings generalize to actual interactions with robots. Future research should address this issue. However, if politeness has a very strong effect on participants' responses in laboratory experiments with images, it should have an even stronger effect when people face robots in real life, where they face the possibility of actual confrontations with the robots.

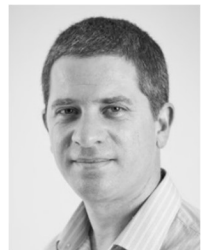

**Ohad Inbar** received the B.Sc. degree in electrical engineering from Ben-Gurion University of the Negev, Beer-Sheva, Israel, in 1992, the M.Sc. degree in industrial design from Ben-Gurion University of the Negev, in 2001, and the Ph.D. degree in industrial engineering from Ben-Gurion University of the Negev, in 2010.

After a Postdoctoral Fellowship with the Ben-Gurion University of theNegev, he continued to serve as a Researcher and a Lecturer with the Department of Industrial Engineering, Tel Aviv University, Tel Aviv, Israel, and with the Faculty of Computer Science, Technion – Israel Institute of Technology, Haifa, Israel.

Dr. Inbar is the Co-founder of IsraHCI (Israel's Human-Computer Interaction organization). His research deals with the design and of user-interfaces and information visualization.

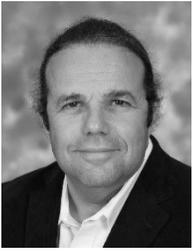

**Joachim Meyer** (M'01-SM'10) received the M.A. degree in psychology from the Ben-Gurion University of the Negev, Beer-Sheva, Israel, in 1986, and the Ph.D. degree in industrial engineering from Ben-Gurion University of the Negev, in 1994.

He held research positions at the Technion – Israel Institute of Technology and MIT, was on the faculty of Ben-Gurion University of the Negev, and is currently a Professor, and was Department Chair, with the Department of Industrial Engineering, Tel Aviv University, Tel Aviv, Israel. Dr. Meyer served for 12 years as an Associate Editor for the IEEE TRANSACTIONS ON SYSTEMS, MAN AND CYBERNETICS: PART A and later the IEEE TRANSACTIONS ON HUMAN-MACHINE SYSTEMS, and is currently on the editorial board of Human Factors.